\documentclass[11pt]{article}

\usepackage[final]{acl}

\usepackage{times}
\usepackage{latexsym}

\usepackage[T1]{fontenc}

\usepackage[utf8]{inputenc}

\usepackage{microtype}

\usepackage{inconsolata}

\usepackage{graphicx}

\usepackage{CJKutf8}
\usepackage{multirow} 
\usepackage{color}
\usepackage[dvipsnames]{xcolor}
\usepackage{enumitem}
\usepackage{CJK}
\usepackage{makecell}
\usepackage{booktabs}
\usepackage{amssymb}
\usepackage{ulem}

\title{CARE: An Explainable Computational Framework for Assessing Client-Perceived Therapeutic Alliance Using Large Language Models}

\author{
 \textbf{Anqi Li\textsuperscript{1,2}},
 \textbf{Chenxiao Wang\textsuperscript{2}},
 \textbf{Yu Lu\textsuperscript{2}},
 \textbf{Renjun Xu\textsuperscript{1}\textsuperscript{*}},
 \textbf{Lizhi Ma\textsuperscript{3}\textsuperscript{*}},
 \textbf{Zhenzhong Lan\textsuperscript{2}\textsuperscript{*}}
\\
 \textsuperscript{1}Zhejiang University
 \\
 \textsuperscript{2}Westlake University
 \\
 \textsuperscript{3}Department of Psychology, Jing Hengyi School of Education, Hangzhou Normal University
\\
 \small{lianqi, lanzhenzhong@westlake.edu.cn; malizhi@hznu.edu.cn}
}

\begin{document}
\begin{CJK*}{UTF8}{gbsn}

\maketitle
\begin{abstract}
Client perceptions of the therapeutic alliance are critical for counseling effectiveness. Accurately capturing these perceptions remains challenging, as traditional post-session questionnaires are burdensome and often delayed, while existing computational approaches produce coarse scores, lack interpretable rationales, and fail to model holistic session context. We present CARE, an LLM-based framework to automatically predict multi-dimensional alliance scores and generate interpretable rationales from counseling transcripts. Built on the CounselingWAI dataset and enriched with 9,516 expert-curated rationales, CARE is fine-tuned using rationale-augmented supervision with the LLaMA-3.1-8B-Instruct backbone. Experiments show that CARE outperforms leading LLMs and substantially reduces the gap between counselor evaluations and client-perceived alliance, achieving over 70\% higher Pearson correlation with client ratings. Rationale-augmented supervision further improves predictive accuracy. CARE also produces high-quality, contextually grounded rationales, validated by both automatic and human evaluations. Applied to real-world Chinese online counseling sessions, CARE uncovers common alliance-building challenges, illustrates how interaction patterns shape alliance development, and provides actionable insights, demonstrating its potential as an AI-assisted tool for supporting mental health care.

\end{abstract}

\section{Introduction}

\begin{figure}[t]
    \centering
    \includegraphics[width=1\linewidth]{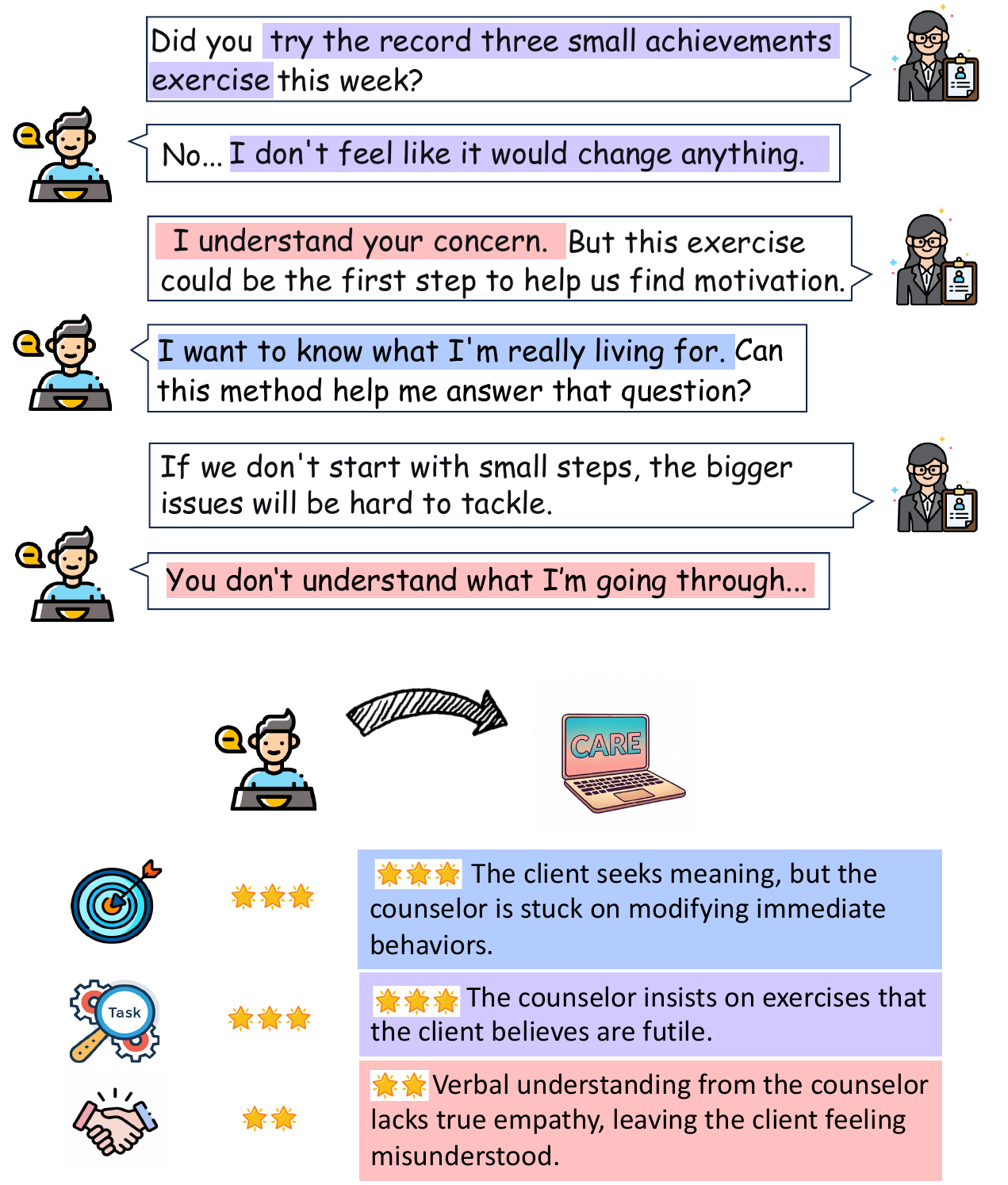}
    \caption{The CARE model predicts client-perceived fine-grained working alliance scores after counseling sessions by identifying and summarizing explanatory reasons from the dialogue, compared to the traditional method where clients only provide scores through questionnaires.}
    \label{fig:first_figure}
\end{figure}

The therapeutic alliance---the collaborative relationship between counselor and client---is one of the strongest predictors of counseling outcomes~\citep{martin2000relation, lambert2001research}. Within this relationship, clients' perceptions are particularly crucial, with studies showing they robustly correlate more strongly with treatment success compared to counselor assessments~\citep{horvath1991relation, piper1991transference}. Moreover, discrepancies in their perceptions may compromise the effectiveness of therapeutic interventions~\citep{horvath2011alliance}.

However, accurately capturing client perspectives remains a major challenge in counseling practice. Traditional methods primarily rely on post-session questionnaires, which can impose burdens on clients and are often completed irregularly or with delays~\citep{goldberg2020machine}. Recent computational approaches have attempted to automatically assess therapeutic alliance from counseling transcripts~\citep{martinez2019personae, goldberg2020machine, lin2023deep}. Yet these approaches face three key limitations: they typically generate global alliance scores while overlooking the multidimensional nature of the therapeutic relationship~\citep{martinez2019personae, goldberg2020machine}; provide only numerical scores without explanatory rationales~\citep{ryu2021natural, goldberg2020machine}; and analyze individual conversation turns in isolation rather than capturing holistic therapeutic context~\citep{lin2023deep}. These gaps restrict the practical utility of AI-driven assessments for guiding counseling practice.

To address these limitations, we introduce CARE\footnote{\textbf{C}lient-perceived \textbf{A}lliance \textbf{R}elationship \textbf{E}valuator}, an LLM-based framework that predicts fine-grained dimensions of client-perceived therapeutic alliance while generating interpretable rationales. CARE is developed under a rationale-augmented supervision paradigm, which injects expert knowledge into the learning process, and incorporates multi-dimensional reasoning to capture associations between session context and specific alliance elements. To support this approach, we enrich the CounselingWAI dataset~\citep{li2024automatic}---originally comprising Chinese text-based counseling sessions with post-session client ratings across three core dimensions: shared goals, coordinated pathways, and emotional attachment---by adding 9,516 expert-annotated rationales linking each dialogue to corresponding alliance dimensions.

CARE is fine-tuned on the LLaMA-3.1-8B-Instruct architecture. Experiments show that CARE significantly bridges the gap where human counselors (who served as practitioners in these counseling sessions) in accurately
gauging the therapeutic alliance from the client's
perspective, achieving over 70\% higher Pearson correlation across all dimensions. It also surpasses leading general-purpose LLMs (e.g., GPT-4o and DeepSeek-R1) across multiple metrics, including correlation, predictive precision, and stability. Further analyses demonstrate that rationale-augmented supervision provides additional gains beyond score-only fine-tuning, and replication on the Qwen2.5-7B model confirms the robustness and generalizability of this approach. Beyond predictive performance, CARE generates high-quality, contextually grounded rationales that align closely with expert references while reflecting human-like reasoning. Human evaluations further validate that these rationales are well-grounded, dimension-specific, and informative.

We then apply the CARE model to a larger dataset of real-world Chinese online counseling conversations. The results reveal common challenges in alliance formation, demonstrate how counselor-client interactions influenced the development of therapeutic alliance, and offer actionable insights for optimizing intervention strategies, highlighting CARE’s potential as an AI-assisted tool for supporting mental health care.

\textbf{The primary contributions of this work are summarized as follows:}

\noindent(1) Dataset enrichment: We augment CounselingWAI with 9,516 expert-annotated rationales linking conversational evidence to client alliance ratings.

\noindent(2) Rationale-augmented framework: We develop CARE, an LLM-based model that predicts fine-grained alliance dimensions while generating interpretable rationales.

\noindent(3) Empirical validation: CARE aligns closely with client perceptions, outperforming human counselors and leading LLMs, with replication across multiple architectures.

\noindent(4) Practical insights: CARE's rationales reveal common alliance-building challenges and patterns in interactions, providing actionable guidance for counseling practice.

\section{Related Work}
\label{sec:related_work}

\paragraph{Computational Alliance Assessment.} A growing body of research has explored the use of NLP techniques to assess the therapeutic alliance from counseling conversations~\citep{goldberg2020machine}. Early studies primarily adopted feature-based machine learning approaches, which represent entire sessions using handcrafted linguistic or statistical features and apply traditional classifiers to predict a global alliance score~\citep{goldberg2020machine, ryu2021natural, martinez2019personae}. While these methods provide a estimation of relational quality, they fail to capture the contextual dynamics and interactive patterns that are central to the counselor–client relationship, and they overlook the multidimensional components of the alliance.

Another line of work employed embedding-based similarity approaches, where semantic embeddings are computed for individual dialogue turns and compared with textual descriptions of questionnaire items to estimate alliance strength~\citep{lin-etal-2024-working, lin2025compass}. Although this paradigm aligns model outputs with theoretical constructs of alliance, it treats conversational turns independently and neglects the global therapeutic context. Moreover, the resulting scores are black-box, offering limited transparency and explainability.

In contrast, our work proposes an explainable computational framework that aligns closely with client-perceived multidimensional alliance while generating interpretable rationales grounded in counseling dialogue.


\paragraph{LLMs in Mental Health Conversation Analysis.}
With the rise of LLMs' advanced text understanding and reasoning capabilities, researchers have increasingly turned to these models for mental health-related analysis based on conversational data~\citep{ji2023rethinking, adhikary2024exploring, chiu2024computational}. Many studies focus on leveraging LLMs to detect mental health conditions, such as anxiety, depression and suicide ideation~\citep{lamichhane2023evaluation, yang2023interpretablementalhealthanalysis, xu2024mentalLLM,yang2023interpretablementalhealthanalysis} and predict the Big Five personality traits~\citep{yan2024predictingbigpersonalitytraits, amin2023will}. 

Some studies explore the use of LLMs to evaluate counseling conversations~\citep{lee-etal-2024-towards, li2024automatic, wang2024towards, lin2025compass}. For example, \citet{lee-etal-2024-towards} use GPT models to classify overall session quality as positive, neutral, or negative, while others assess therapeutic alliance through dialogue~\citep{li2024automatic, wang2024towards}. Notably, \citet{li2024automatic} propose a prompting-based framework with Chain-of-Thought reasoning to approximate expert-rated alliance. In contrast, we model client-perceived alliance which is more directly linked to treatment outcomes, and move beyond prompting by adopting a rationale-augmented supervision paradigm. By leveraging 9,516 expert-curated rationales, we fine-tune a specialized model that internalizes domain knowledge rather than relying solely on general-purpose LLM capabilities.





\section{Measurement of Therapeutic Alliance}
\label{sec:definition}

The therapeutic alliance is broadly recognized as a collaborative element of the client-counselor relationship~\citep{bordin1979generalizability, ardito2011therapeutic}. This multifaceted concept, which integrates both cognitive and emotional interactions, is generally characterized by three components: (a) mutual agreement on the goals of therapy (\textbf{Goal}); (b) a shared understanding that the therapeutic tasks will effectively address clients’ specific concerns (\textbf{Task}); and (c) the strength of the interpersonal bond between clients and counselors (\textbf{Bond})~\citep{bordin1979generalizability}. Each dimension captures a distinct yet complementary aspect of the alliance, enabling a comprehensive assessment of the therapeutic process.

In both clinical counseling and research, the short revised Working Alliance Inventory (WAI) is among the most widely used measures of these three alliance dimension. The inventory is available in two versions—one for counselors and one for clients—that are nearly identical, differing only in the grammatical subject. It consists of 12 items, with four items devoted to each dimension of the therapeutic alliance. Each item is rated on a 5-point Likert scale: 1 = Seldom; 2 = Sometimes; 3 = Fairly Often; 4 = Very Often; 5 = Always. Dimension scores are derived by averaging the four items within each subscale. The reliability and validity of this inventory have been well-established across various forms of psychotherapy~\citep{hatcher2006development, munder2010working}.

\section{Data Collection}
\label{sec:data}

\subsection{Data Source}
\label{subsec: data_source}
The counseling conversations and corresponding therapeutic alliance scores used in this study are drawn from the research-restricted CounselingWAI dataset~\citep{li2024automatic}. This corpus comprises text-based, 50-minute counseling sessions conducted between professional counselors and real clients on an online Chinese psycho-counseling platform. Following each session, both clients and counselors completed their respective versions of the short revised Working Alliance Inventory. For this study, we used a subset comprising 793 counseling conversations from 82 clients, each paired with client-reported working alliance scores. Among these, 728 sessions also include counselor-reported alliance scores. The distributions of client ratings across each alliance dimension are shown in Appendix~\ref{appendix: data_analysis}.

\subsection{Rationale Augmentation}

The original dataset contained client ratings of the working alliance as numerical scores only (see \S\ref{subsec: data_source}), which provided no explanation for the evaluations. To enrich this data with interpretative evidence, we engaged two licensed counseling experts who also serve as clinical supervisors skilled in providing feedback and guidance to counselors. Their task was to identify supporting evidence from the counseling conversations for each of the 12 item–score pairs in the questionnaire.

Drawing on their counseling and supervision expertise, the experts concluded that high-quality rationales should combine a holistic, session-level analysis of counselor–client interactions with turn-level identification of critical exchanges that likely influenced clients' ratings. This integrated approach captures overall session dynamics while reducing bias from overemphasizing isolated turns.

To facilitate rationale construction, we adopted a model-in-the-loop strategy, which has been shown to enhance annotation quality~\citep{li-etal-2023-coannotating}. In our setup, GPT-4o was prompted to generate draft rationales using counseling dialogues alongside corresponding items and client-provided ground-truth ratings. These drafts were then reviewed and refined by one of the expert annotators to ensure both fidelity to the original dialogue and consistency with the clients' perceived scores. For instance, when a client assigned a low rating to the item "I have gained a clearer understanding of what changes I can make" within the task-setting dimension, the finalized rationale read: "The client expressed confusion about communication difficulties, but the counselor primarily facilitated self-expression without offering actionable suggestions. The client explicitly remarked, \textit{I still don't know what to do; the counseling time was not worth the cost.}" Through this process, we developed 9,516 explanations (793 dialogues × 12 items) for rationale augmentation.


\subsection{Data Quality}
\label{sec:data_quality}

To ensure the quality of the generated rationales, we implemented a cross-evaluation protocol in which each expert assessed the complete set of rationales corresponding to 100 counseling conversations authored by their counterpart. This process yielded a total of 2,400 rationale evaluations, representing approximately 25\% of the overall dataset. The assessment was conducted using a 5-point Likert scale along three dimensions, with a score of 3 set as the acceptability threshold. The evaluation criteria comprised: (1) \textit{Faithfulness} - how accurately the rationale reflected both the dialogue content and the client's actual rating; (2) \textit{Relevance} - how directly the rationale addressed the specific aspect being evaluated; and (3) \textit{Informativeness} - whether the rationale provided context-specific rather than generic statements. Results demonstrated high quality across all aspects, with mean scores of $4.82_{0.39}$ for \textit{Faithfulness}, $4.50_{0.65}$ for \textit{Relevance}, and $4.20_{0.55}$ for \textit{Informativeness}. 

We further examine lexical patterns for each dimension's rationales by computing the log odds ratio with an informative Dirichlet prior~\citep{monroe2008fightin}, comparing all unigrams in the rationales of one dimension against the other two. Rationales for each dimension were significantly associated with specific key phrases (e.g., \textit{Goal} with "establish," \textit{Task} with "method," \textit{Bond} with "support") (see Appendix~\ref{appendix: data_analysis}).

\subsection{Privacy and Ethics}
All personal identifiers in the original CounselingWAI dialogue data have been thoroughly anonymized to ensure participant privacy. In accordance with CounselingWAI's data sharing policy, the data will only be accessible to researchers who have obtained proper authorization from CounselingWAI. This work is intended to serve as an auxiliary tool for counselors to better understand their clients' perception of the working alliance. Its outputs should be adopted with caution in practice and do not provide any treatment recommendations or diagnostic claims.

\section{CARE Model}
\label{sec:experiment}

Leveraging the collected dataset, we developed CARE, an automatic LLM-based model for predicting client-perceived working alliance scores from text-based conversations, while simultaneously generating context-sensitive explanations.


\subsection{Task Definition}
Formally, we define the task of therapeutic relationship evaluation as follows: given a counseling conversation and each corresponding measurement item, the model predicts the client's rating and identifies supporting evidence from the dialogue. For each dimension, the predicted score is computed as the average of the predicted ratings across its four representative items. Model performance is then evaluated per dimension by calculating the Pearson correlation ($r$), Spearman correlation ($\rho$), and mean squared error (MSE) with respect to the client's self-reported scores, following evaluation protocols widely adopted in prior work~\citep{shick2007meta, bachelor2000participants, goldberg2020machine, lalk2024alliance}. For model-generated rationales, we further assess their quality through both automatic and human evaluations. Automatic metrics include BLEU, ROUGE-1, ROUGE-L, and BERTScore, which quantify textual similarity to expert-written references. Additionally, human experts rate the rationales along three dimensions (faithfulness, relevance, and informativeness, following the procedures described in \S~\ref{sec:data_quality}).


\subsection{Data Preparation}
To obtain robust performance estimates and reduce bias from a single train–test split, we adopted a 5-fold cross-validation. The dataset was divided into five mutually exclusive subsets, with stratified sampling applied based on the score distributions of the three working alliance dimensions. This ensured consistent score distributions across folds. To prevent data leakage, all sessions from the same client were assigned exclusively to either the training or validation set. This design minimizes the risk of client-specific memorization and better reflects real-world deployment, where the model must generalize to unseen clients. 

As summarized in Table~\ref{tab:data_split}, we report for each fold the number of counseling sessions and clients, along with the mean and standard deviation of the goal, task, and bond scores. During training, models were fine-tuned on four folds and validated on the hold-out fold, with the optimal checkpoint selected based on Pearson correlation on the validation set. For prompt-based approaches, models were directly applied to each validation fold. Final performance is reported as the average and standard deviation across all five folds.

\begin{table}[]
\centering
\scalebox{0.8}{
\begin{tabular}{cccccc}
\toprule
\textbf{Fold} & \textbf{\#Client} & \textbf{\#Session} & \textbf{Goal} & \textbf{Task} & \textbf{Bond} \\
\midrule
1             & 15                & 147                & $3.74_{1.07}$       & $3.52_{1.15}$       & $3.74_{1.06}$       \\
2             & 16                & 154                & $3.71_{1.12}$       & $3.51_{1.06}$       & $4.05_{0.91}$       \\
3             & 16                & 157                & $4.15_{0.81}$       & $3.97_{0.86}$       & $4.31_{0.80}$       \\
4             & 18                & 171                & $4.11_{0.93}$       & $3.87_{0.98}$       & $4.22_{0.87}$       \\
5             & 17                & 164                & $3.69_{1.05}$       & $3.36_{1.10}$       & $3.99_{0.93}$    \\ 
Total & 82 & 793 & $3.89_{1.02}$ & $3.65_{1.06}$ & $4.07_{0.93}$ \\
\bottomrule
\end{tabular}}
\caption{Summary statistics for each fold, including the number of clients, number of sessions, and the mean with standard deviation of goal, task, and bond scores.}
\label{tab:data_split}
\end{table}




\begin{table*}[]
\centering
\scalebox{0.68}{
\begin{tabular}{c|ccc|ccc|ccc}
\toprule
\multirow{2}{*}{\textbf{Model Name}} & \multicolumn{3}{c}{\textbf{Goal}}                                        & \multicolumn{3}{c}{\textbf{Task}}                                        & \multicolumn{3}{c}{\textbf{Bond}}                                        \\
& \textbf{Pearson}$\uparrow$       & \textbf{Spearman}$\uparrow$      & \textbf{MSE}$\downarrow$           & \textbf{Pearson}$\uparrow$       & \textbf{Spearman}$\uparrow$      & \textbf{MSE}$\downarrow$           & \textbf{Pearson}$\uparrow$       & \textbf{Spearman}$\uparrow$      & \textbf{MSE}$\downarrow$           \\
\hline
\textbf{Human Counselor}             & 0.30                  &   0.27                     &  1.36                      & 0.30                   &  0.28                      &     1.61                   & 0.22                   &    0.21                    &      1.24                  \\
\hline
\textbf{GPT-3.5-Turbo}               & $0.27_{0.11}$          & $0.24_{0.11}$          & $2.10_{0.37}$          & $0.30_{0.11}$          & $0.29_{0.11}$          & $1.16_{0.13}$          & $0.30_{0.09}$          & $0.28_{0.10}$          & $1.89_{0.30}$          \\
\textbf{GPT-4o-mini}                 & $0.35_{0.11}$          & $0.31_{0.12}$          & $1.27_{0.19}$          & $0.36_{0.11}$          & $0.34_{0.14}$          & $1.11_{0.15}$          & $0.34_{0.12}$          & $0.31_{0.12}$          & $0.90_{0.04}$          \\
\textbf{GPT-4o}                      & $0.43_{0.07}$          & $0.40_{0.09}$          & $2.03_{0.40}$          & $0.48_{0.08}$          & $0.49_{0.10}$          & $1.62_{0.37}$          & $0.37_{0.15}$          & $0.35_{0.14}$          & $0.97_{0.12}$          \\
\textbf{Claude-3-Sonnet}             & $0.35_{0.13}$          & $0.36_{0.13}$          & $2.09_{0.39}$          & $0.43_{0.13}$          & $0.41_{0.14}$          & $1.47_{0.29}$          & $0.29_{0.10}$          & $0.23_{0.09}$          & $1.51_{0.17}$          \\
\hline
\textbf{DeepSeek-R1}                 & $0.42_{0.14}$          & $0.39_{0.14}$          & $2.36_{0.55}$          & $0.49_{0.12}$          & $0.48_{0.13}$          & $1.17_{0.26}$          & $0.40_{0.17}$          & $0.37_{0.14}$          & $0.77_{0.21}$          \\
\textbf{Qwen2.5-7B-Instruct}         & $0.22_{0.06}$          & $0.21_{0.05}$          & $2.63_{0.26}$          & $0.28_{0.08}$          & $0.27_{0.08}$          & $1.76_{0.18}$          & $0.25_{0.12}$          & $0.23_{0.09}$          & $1.25_{0.16}$          \\
\textbf{Qwen2.5-14B-Instruct}        & $0.30_{0.11}$          & $0.28_{0.11}$          & $2.17_{0.22}$          & $0.37_{0.08}$          & $0.35_{0.08}$          & $1.76_{0.12}$          & $0.30_{0.08}$          & $0.27_{0.06}$          & $2.40_{0.21}$          \\
\textbf{Qwen2.5-32B-Instruct}        & $0.32_{0.08}$          & $0.30_{0.09}$          & $2.51_{0.33}$          & $0.36_{0.07}$          & $0.35_{0.08}$          & $1.83_{0.24}$          & $0.27_{0.09}$          & $0.25_{0.07}$          & $2.37_{0.15}$          \\
\textbf{Qwen2.5-72B-Instruct}        & $0.30_{0.10}$          & $0.29_{0.09}$          & $1.95_{0.22}$          & $0.34_{0.09}$          & $0.33_{0.08}$          & $1.62_{0.13}$          & $0.27_{0.10}$          & $0.24_{0.08}$          & $2.06_{0.23}$          \\
\textbf{Llama-3.1-8B-Instruct}       & $0.24_{0.10}$          & $0.24_{0.10}$          & $2.98_{0.54}$          & $0.36_{0.08}$          & $0.36_{0.09}$          & $2.33_{0.47}$          & $0.26_{0.14}$          & $0.27_{0.12}$          & $3.31_{0.67}$          \\
\textbf{Llama-3.1-70B-Instruct}      & $0.40_{0.12}$          & $0.38_{0.12}$          & $2.20_{0.42}$          & $0.45_{0.11}$          & $0.43_{0.12}$          & $1.32_{0.24}$          & $0.33_{0.11}$          & $0.30_{0.09}$          & $0.92_{0.10}$          \\
\hline
\textbf{CARE (Our Model)}            & $\mathbf{0.52_{0.06}}$ & $\mathbf{0.50_{0.07}}$ & $\mathbf{1.00_{0.16}}$ & $\mathbf{0.50_{0.08}}$ & $\mathbf{0.49_{0.08}}$ & $\mathbf{1.05_{0.04}}$ & $\mathbf{0.46_{0.05}}$ & $\mathbf{0.41_{0.04}}$ & $\mathbf{0.70_{0.10}}$ \\
\bottomrule
\end{tabular}}
\caption{Performance of different models across the three alliance dimensions, including human counselors ratings, baseline models, and the proposed CARE model, benchmarked against clients’ self-reported ratings. Evaluation metrics include the Pearson's $r$, Spearman's $\rho$, and MSE. $\uparrow$/$\downarrow$ indicates that higher/lower values are better. The best results in each column are shown in bold.}
\label{tab:main_results}
\end{table*}

\subsection{Experimental Setup}

We trained our CARE model using LLaMA-3.1-8B-Instruct as the backbone, a widely adopted open-source instruction-tuned model chosen to balance performance and deployment efficiency. Supervised full-parameter fine-tuning was conducted on the training data for 10 epochs with a learning rate of $5 \times 10^{-7}$ and a fixed random seed of 123. During inference, the temperature and nucleus sampling parameters were set to 0 and 1.0, respectively, to ensure deterministic outputs.

Model training was conducted using LLaMA-Factory~\citep{zheng2024llamafactory}, and all experiments were run on eight NVIDIA A100 (80GB) GPUs. Additional implementation details, including the template prompt, are provided in Appendix~\ref{template_prompt}.

\subsection{Baselines}

To evaluate the performance of our proposed CARE model in estimating clients’ perceived therapeutic alliance, we compared it against several representative baselines, including human counselors and large language models (LLMs).


\noindent\textbf{Human Counselors.}
We incorporated the alliance ratings provided by the counselors themselves, as originally collected in the dataset (see \S\ref{subsec: data_source}). Human performance was computed across all 729 sessions for which both counselor and client alliance ratings were available.


\noindent\textbf{LLMs.}
We further examined the performance of advanced open- and closed-source LLMs in a zero-shot setting. The closed-source models include ChatGPT~\citep{chatgpt}, GPT-4o-mini~\citep{gpt-4o-mini}, GPT-4o~\citep{2023gpt4}, and Claude-3-Sonnet~\citep{Anthropic2024}. The open-source models include the Qwen2.5-Instruct series~\citep{qwen2.5} (7B, 14B, 32B, and 72B), the LLaMA-3.1-Instruct series~\citep{llama3modelcard} (8B and 70B), and DeepSeek-R1~\citep{deepseekai2025deepseekr1}.

\subsection{Additional Experiments}

\paragraph{Ablation Study on Augmented Rationales.} 
To examine the contribution of the augmented rationales to model performance, we conducted an ablation study in which the LLaMA-3.1-8B-Instruct model was trained on the dataset containing only client-provided ratings, with all rationale annotations removed.

\paragraph{Generalization Across Different Architectures.}
To further assess the robustness and generalizability of the proposed dataset, we replicated the experiments using the Qwen2.5-7B-Instruct model, following the same fine-tuning procedure as employed for our CARE model and its ablated variant.





\begin{table}[]
\centering
\scalebox{0.85}{
\begin{tabular}{ccccc}
\toprule
\multicolumn{2}{c}{\textbf{Metrics}}                           & \textbf{Goal} & \textbf{Task} & \textbf{Bond} \\ \hline 
\multirow{4}{*}{\rotatebox{90}{\textbf{Automatic}}} & \textbf{BLEU}            & $0.22_{0.01}$ & $0.23_{0.00}$ & $0.28_{0.01}$ \\
                           & \textbf{ROUGE-1}         & $0.56_{0.00}$ & $0.55_{0.00}$ & $0.57_{0.01}$ \\
                           & \textbf{ROUGE-L}         & $0.42_{0.00}$ & $0.42_{0.00}$ & $0.47_{0.01}$ \\
                           & \textbf{BERTScore}       & $0.79_{0.00}$ & $0.79_{0.00}$ & $0.81_{0.00}$ \\ \hline
\multirow{3}{*}{\rotatebox{90}{\textbf{Human}}}     & \textbf{Faithfulness}    & $4.77_{0.08}$ & $4.87_{0.04}$ & $4.81_{0.07}$ \\
                           & \textbf{Relevance}       & $4.50_{0.12}$ & $4.73_{0.03}$ & $4.78_{0.11}$ \\
                           & \textbf{Informativeness} & $4.34_{0.09}$ & $4.45_{0.11}$ & $4.27_{0.10}$ \\ \bottomrule
\end{tabular}}
\caption{Comparison of CARE-generated rationales and expert-written rationales in terms of automatic evaluation metrics (BLEU, ROUGE-1, ROUGE-L, and BERTScore) and human assessment dimensions (Faithfulness, Relevance, and Informativeness).}
\label{tab:generation_metrics}
\end{table}

\begin{table*}[]
\centering
\scalebox{0.65}{
\begin{tabular}{cc|ccc|ccc|ccc}
\toprule
\multirow{2}{*}{\textbf{Architectures}}   & \multirow{2}{*}{\textbf{Rationale}} & \multicolumn{3}{c}{\textbf{Goal}}                                        & \multicolumn{3}{c}{\textbf{Task}}                                        & \multicolumn{3}{c}{\textbf{Bond}}                                        \\
&                                      & \textbf{Pearson}$\uparrow$       & \textbf{Spearman}$\uparrow$      & \textbf{MSE}$\downarrow$          & \textbf{Pearson}$\uparrow$       & \textbf{Spearman}$\uparrow$      & \textbf{MSE}$\downarrow$            & \textbf{Pearson}$\uparrow$       & \textbf{Spearman}$\uparrow$      & \textbf{MSE}$\downarrow$            \\
\hline
\multirow{2}{*}{\textbf{Qwen2.5-7B-Instruct}}   &  $\times$      & $0.37_{0.12}$          & $0.36_{0.14}$          & $1.55_{0.20}$          & $0.38_{0.13}$          & $0.37_{0.13}$          & $1.60_{0.17}$          & $0.34_{0.12}$          & $0.31_{0.09}$          &    $1.44_{0.25}$         \\
& $\checkmark$                                   & $0.42_{0.08}$          & $0.40_{0.10}$          & $1.49_{0.17}$          & $0.41_{0.09}$          & $0.41_{0.11}$          & $1.58_{0.10}$          & $0.36_{0.09}$          & $0.33_{0.06}$          & $1.37_{0.24}$        \\
                                       \hline
\multirow{2}{*}{\textbf{Llama-3.1-8B-Instruct}} & $\times$                                   & $0.45_{0.09}$          & $0.43_{0.11}$          & $1.10_{0.14}$          & $0.45_{0.11}$          & $0.44_{0.10}$          & $1.18_{0.11}$          & $0.44_{0.07}$          & $0.39_{0.07}$          & $0.92_{0.17}$          \\
& $\checkmark$                                 & $\mathbf{0.52_{0.06}}$ & $\mathbf{0.50_{0.07}}$ & $\mathbf{1.00_{0.16}}$ & $\mathbf{0.50_{0.08}}$ & $\mathbf{0.49_{0.08}}$ & $\mathbf{1.05_{0.04}}$ & $\mathbf{0.46_{0.05}}$ & $\mathbf{0.41_{0.04}}$ & $\mathbf{0.70_{0.10}}$ \\
\bottomrule
\end{tabular}}
\caption{Results of generalization and ablation studies for fine-tuning with versus without rationales, using Qwen2.5-7B-Instruct and LLaMA-3.1-8B-Instruct as backbones. Evaluation metrics include the Pearson's $r$, Spearman's $\rho$, and MSE. $\uparrow$/$\downarrow$ indicates that higher/lower values are better. The best results in each column are shown in bold.}
\label{tab:additional_exp}
\end{table*}

\section{Results and Analysis}

\subsection{Main Results}

Table~\ref{tab:main_results} presents the performance of different models across the three alliance dimensions, benchmarked against clients’ self-reported ratings under all experimental settings. The comparison includes ratings from human counselors, all baseline models, and our proposed CARE model. From these results, we have several findings:

\paragraph{\textbf{CARE significantly bridges the gap where human counselors may struggle to accurately gauge the therapeutic alliance from the client's perspective.}} Human counselors exhibited only moderate alignment with client self-reports, with Pearson correlations ranging from 0.22 to 0.30 across the Goal, Task, and Bond dimensions. This is consistent with prior findings that counselors often struggle to accurately assess the counseling quality from their clients’ perspectives~\citep{hatcher1995patients, shick2007meta}, highlighting the inherent difficulty of this task. In contrast, the CARE model achieved substantially higher correlations of 0.52 (Goal), 0.50 (Task), and 0.46 (Bond), corresponding to improvements of approximately 73\%, 67\%, and 109\% in Pearson correlation for the respective dimensions. 

These results suggest that the CARE model could serve as a reliable tool for providing quantitative feedback to counselors, helping them better calibrate their perception of the therapeutic alliance against clients’ reported experiences.


\paragraph{\textbf{CARE outperforms even the most powerful general-purpose LLMs in correlation, predictive precision, and stability.}}

Among zero-shot prompting models, GPT-4o and DeepSeek-R1 exhibited the strongest performance. Compared to these top competitors, CARE consistently outperformed across all three alliance dimensions, with notable gains on the Goal and Bond dimensions. Specifically, CARE improved Pearson and Spearman correlations by 21\% and 25\% on the Goal dimension, and by 15\% and 11\% on the Bond dimension.

In terms of predictive precision, CARE achieved the lowest MSE values across all three dimensions (Goal: 1.00, Task: 1.05, Bond: 0.70). On the Goal dimension, CARE’s MSE was 51\% lower than GPT-4o (2.03) and 58\% lower than DeepSeek-R1 (2.36), indicating that its predictions were not only highly correlated with client ratings but also numerically closer to the ground-truth scores. Furthermore, CARE exhibited remarkably low standard deviations compared to the top competitors, ensuring stable and reliable performance in real-world applications.

Beyond predictive metrics, the rationales generated by CARE also show high semantic alignment with expert-written references, despite lexical variations (see Table~\ref{tab:generation_metrics}). This finding reflects the model's ability to construct human-like reasoning chains from dialogue rather than merely replicating reference expressions. Human evaluation further confirms the quality of generated rationales, demonstrating strong grounding in session content, accurate focus on dimension-specific cues and adequate coverage of contextual information.


\paragraph{\textbf{CARE transformed its backbone into a highly accurate, stable, and reliable model in predicting therapeutic alliance.}}
The backbone of CARE, LLaMA-3.1-8B-Instruct, showed relatively poor performance under zero-shot prompting, with correlations as low as 0.24 (Goal) and 0.26 (Bond), and MSE values exceeding 2 across dimensions. After fine-tuning on our proposed dataset, CARE achieved substantial improvements, corresponding to 117\%, 47\%, and 77\% increases in Pearson correlation on the Goal, Task, and Bond dimensions, respectively. MSE values dropped to approximately 1 across all dimensions, with the Bond dimension decreasing from 3.31 to 0.70 (a 79\% reduction). Furthermore, the high standard deviations of the base model were drastically reduced, demonstrating that fine-tuning specialized the model for this task, yielding consistent and reliable predictions.



\subsection{Additional Results}

Table~\ref{tab:additional_exp} summarizes the ablation study evaluating the effect of rationale augmentation and its generalizability across model architectures.

\paragraph{\textbf{Rationale-augmented supervision enhanced CARE's predictive capability.}}

Fine-tuning the CARE model with rationale-augmented supervision yielded consistent and significant gains across all therapeutic alliance dimensions compared to training without rationales. Correlation metrics improved notably—by approximately 16\% on the Goal dimension and 11\% on the Task dimension—while predictive errors (MSE) were markedly reduced, exemplified by a reduction from 0.92 to 0.70 on the Bond dimension. These results highlight the efficacy of incorporating expert-generated textual rationales during training, enabling the model to internalize human evaluative reasoning and attend to high-quality cues for more reliable alliance assessment.

\paragraph{CARE's rationale-augmented training generalized across model architectures.}

Consistent with the findings on LLaMA-3.1-8B-Instruct, the Qwen2.5-7B-Instruct model exhibited a similar performance pattern, showing a substantial improvement from score-based fine-tuning, followed by further gains with rationale-augmented training. This consistency demonstrates that the benefits of rationale supervision are architecture-agnostic, validating the robustness and high quality of our proposed dataset.

\subsection{Case Study}

To better understand the limitations of the CARE model, we examined cases where its predictions deviated from client-reported ratings (see Table~\ref{tab:case}) and analyzed model-generated explanations to identify the underlying causes.

Our analysis revealed that the model tends to prioritize verbal content while overlooking important behavioral cues. For instance, in the first case, the client arrived late and departed early, suggesting limited respect and weak emotional engagement. However, the model focused exclusively on the superficially polite language and inferred a high level of respect. Moreover, the model may overemphasize some positive or negative statements while neglecting the broader contextual dynamics. In the second case, although the client initially acknowledged the benefits of previous counseling, they later provided minimal responses and disengaged from session activities. Yet, the model weighted the positive statements disproportionately, failing to capture the deteriorating therapeutic alliance. 

These findings align closely with insights emphasized by experts during our rationale augmentation process—that accurate assessment of therapeutic alliance requires integrative analysis of both specific verbal exchanges and the overall interaction pattern. Enhancing this integrative reasoning capability represents a crucial direction for future model refinement.




\section{Insights based on LLM Predictions}
\label{sec:insights}

We use the CARE model to explore therapeutic alliance in text-based psychological counseling. This model predicted client-perceived alliance across 2,236 sessions from the ClientBehavior dataset~\citep{li-etal-2023-understanding}, including a subset of 300 sessions annotated with utterance-level counselor strategies and client reactions. Our analysis examines counselors’ varying abilities to build alliances, how interaction patterns influence alliance development, and actionable directions for improving practice.

\begin{figure}
    \centering
    \includegraphics[width=1.02\linewidth]{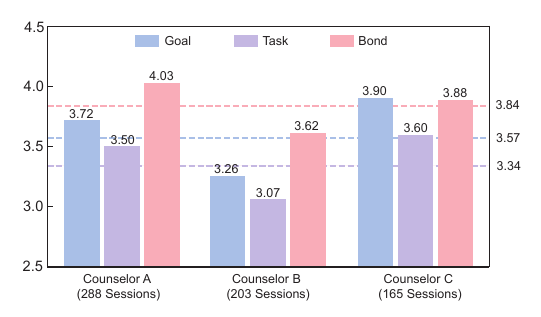}
    \caption{Working alliance scores by dimension for the three most active counselors; dashed lines indicate overall counselor averages.}
    \label{fig:counselor_wai}
\end{figure}

\paragraph{Counselors' Varying Abilities.} Overall, counselors formed a generally solid therapeutic alliance, though substantial room for improvement remains (see Figure~\ref{fig:counselor_wai}). Collaboration on counseling tasks, a core mechanism for behavioral change, emerged as the weakest dimension.

Analysis of the three counselors with the highest session volumes (≥165 sessions) revealed distinct alliance profiles. Counselor B scored below the overall mean across all dimensions, indicating ongoing difficulties in aligning with clients, especially on goals and tasks. In contrast, Counselor A showed excelled in building emotional trust (bond > 4), whereas Counselor C performed best in establishing shared goals. Since these dimensions ideally develop together~\citep{bordin1979generalizability}, these patterns highlight the need for counselors to improve their weaker areas.

\paragraph{Interaction Patterns Affecting Alliance.} We analyzed micro-level interaction patterns in the 300 annotated sessions using multiple regression, with counselor–client patterns as predictors and session-level alliance scores as outcomes.

Across all dimensions, the combination of counselor \textit{Supporting} strategies and clients’ \textit{Negative} responses was linked to lower alliance scores, strongest for the bond dimension ($\beta$ = −6.36, $p$ < 0.001). In contrast, counselor \textit{Challenging} strategies paired with clients’ \textit{Positive} responses predicted higher alliance scores, particularly on the task dimension ($\beta$ = 2.64, $p$ < 0.001). These findings underscore the bidirectional nature of counseling interactions and the importance of adapt strategies in response to client feedback to strengthen the alliance.

\paragraph{Actionable Insights for Counselors.}
Focusing on Counselor B, who struggled to establish strong alliances, model explanations for sessions with low alliance scores (≤ 2) revealed clear interaction patterns.

The counselor often struggled to set concrete goals with clients, with discussions drifting into open-ended exploration. Counseling activities (e.g., overuse of "mm-hmm" responses) were misaligned with goals, and suggestions were vague and not actionable. Responses appeared mechanical, empathy was limited, and attempts to repair the relationship were largely ineffective when clients expressed dissatisfaction. These observations suggest that Counselor B would benefit from improving goal clarity, aligning tasks with goals, and providing more responsive emotional feedback.

Building on these insights, we aim to integrate LLM-based real-time evaluation and feedback into counseling practice to support counselors in cultivating deeper connections with their clients in future work.

\section{Conclusion}
\label{sec:conclusion}

We augment counseling dialogues with expert-curated rationales aligned with client-reported working alliance scores and introduce a novel LLM framework for assessing multidimensional therapeutic alliance with interpretable evidence. Experiments show that a smaller fine-tuned LLM with rationale-based supervision aligns with client perceptions more closely than both human counselors and larger general-purpose LLMs. Moreover, CARE demonstrates its potential as an AI-assisted tool for supporting text-based mental health care. Data, code, and models are available at \href{https://anonymous.4open.science/r/CARE---Assessing-Client-WAI-Using-LLMs-6672/}{this URL}.



\bibliography{custom}

@article{tryon2008magnitude,
  title={The magnitude of client and therapist working alliance ratings.},
  author={Tryon, Georgiana Shick and Blackwell, Sasha Collins and Hammel, Elizabeth Felleman},
  journal={Psychotherapy: Theory, Research, Practice, Training},
  volume={45},
  number={4},
  pages={546},
  year={2008},
  publisher={Educational Publishing Foundation}
}

@article{goldberg2020machine,
  title={Machine learning and natural language processing in psychotherapy research: Alliance as example use case.},
  author={Goldberg, Simon B and Flemotomos, Nikolaos and Martinez, Victor R and Tanana, Michael J and Kuo, Patty B and Pace, Brian T and Villatte, Jennifer L and Georgiou, Panayiotis G and Van Epps, Jake and Imel, Zac E and others},
  journal={Journal of counseling psychology},
  volume={67},
  number={4},
  pages={438},
  year={2020},
  publisher={American Psychological Association}
}

@article{shick2007meta,
  title={A meta-analytic examination of client--therapist perspectives of the working alliance},
  author={Shick Tryon, Georgiana and Collins Blackwell, Sasha and Felleman Hammel, Elizabeth},
  journal={Psychotherapy research},
  volume={17},
  number={6},
  pages={629--642},
  year={2007},
  publisher={Taylor \& Francis}
}

@article{hatcher1995patients,
  title={Patients' and therapists' shared and unique views of the therapeutic alliance: an investigation using confirmatory factor analysis in a nested design.},
  author={Hatcher, Robert L and Barends, Alex and Hansell, James and Gutfreund, M Janice},
  journal={Journal of Consulting and Clinical Psychology},
  volume={63},
  number={4},
  pages={636},
  year={1995},
  publisher={American Psychological Association}
}

@misc{Anthropic2024,
	author = {Anthropic},
	title = {{T}he claude 3 model family: {O}pus, sonnet, haiku},
	howpublished = {\url{https://api. semanticscholar.org/CorpusID:268232499}},
	year = {2024},
	note = {[Accessed 16-04-2024]},
}

@article{2023gpt4,
  title={Gpt-4 technical report},
  author={OpenAI},
  journal={arXiv preprint arXiv:2303.08774},
  year={2023}
}

@misc{chatgpt,
  title={ChatGPT},
  author={OpenAI},
  year={2023},
  url={https://chat.openai.com.chat.}}

@article{llama3modelcard,
title={Llama 3 Model Card},
author={AI@Meta},
year={2024},
url = {https://github.com/meta-llama/llama3/blob/main/MODEL_CARD.md},
journal={}
}

@Misc{gpt-4o-mini,
url = {https://openai.com/index/gpt-4o-mini-advancing-cost-efficient-intelligence/},
title = {GPT-4o mini: advancing cost-efficient intelligence},
author = {OpenAI},
year={2024}}

@article{martin2000relation,
  title={Relation of the therapeutic alliance with outcome and other variables: a meta-analytic review.},
  author={Martin, Daniel J and Garske, John P and Davis, M Katherine},
  journal={Journal of consulting and clinical psychology},
  volume={68},
  number={3},
  pages={438},
  year={2000},
  publisher={American Psychological Association}
}

@article{lambert2001research,
  title={Research summary on the therapeutic relationship and psychotherapy outcome.},
  author={Lambert, Michael J and Barley, Dean E},
  journal={Psychotherapy: Theory, research, practice, training},
  volume={38},
  number={4},
  pages={357},
  year={2001},
  publisher={Division of Psychotherapy (29), American Psychological Association}
}

@inproceedings{martinez2019personae,
  title={Identifying therapist and client personae for therapeutic alliance estimation},
  author={Martinez, Victor R and Flemotomos, Nikolaos and Ardulov, Victor and Somandepalli, Krishna and Goldberg, Simon B and Imel, Zac E and Atkins, David C and Narayanan, Shrikanth},
  booktitle={Interspeech},
  volume={2019},
  pages={1901},
  year={2019},
  organization={NIH Public Access}
}

@article{ryu2021natural,
  title={A natural language processing approach to modelling treatment alliance in psychotherapy transcripts},
  author={Ryu, Jihan and Heisig, Stephen and McLaughlin, Caroline and Bortz, Rebeccah and Katz, Michael and Gu, Xiaosi},
  journal={BJPsych Open},
  volume={7},
  number={S1},
  pages={S48--S48},
  year={2021},
  publisher={Cambridge University Press}
}

@inproceedings{lin2023deep,
  title={Deep annotation of therapeutic working alliance in psychotherapy},
  author={Lin, Baihan and Cecchi, Guillermo and Bouneffouf, Djallel},
  booktitle={International workshop on health intelligence},
  pages={193--207},
  year={2023},
  organization={Springer}
}

@article{bordin1979generalizability,
  title={The generalizability of the psychoanalytic concept of the working alliance.},
  author={Bordin, Edward S},
  journal={Psychotherapy: Theory, research \& practice},
  volume={16},
  number={3},
  pages={252},
  year={1979},
  publisher={Division of Psychotherapy (29), American Psychological Association}
}

@article{ardito2011therapeutic,
  title={Therapeutic alliance and outcome of psychotherapy: historical excursus, measurements, and prospects for research},
  author={Ardito, Rita B and Rabellino, Daniela},
  journal={Frontiers in psychology},
  volume={2},
  pages={270},
  year={2011},
  publisher={Frontiers Research Foundation}
}

@article{horvath1993role,
  title={The role of the therapeutic alliance in psychotherapy.},
  author={Horvath, Adam O and Luborsky, Lester},
  journal={Journal of consulting and clinical psychology},
  volume={61},
  number={4},
  pages={561},
  year={1993},
  publisher={American Psychological Association}
}

@misc{yang2023interpretablementalhealthanalysis,
      title={Towards Interpretable Mental Health Analysis with Large Language Models}, 
      author={Kailai Yang and Shaoxiong Ji and Tianlin Zhang and Qianqian Xie and Ziyan Kuang and Sophia Ananiadou},
      year={2023},
      eprint={2304.03347},
      archivePrefix={arXiv},
      primaryClass={cs.CL},
      url={https://arxiv.org/abs/2304.03347}, 
}

@misc{yan2024predictingbigpersonalitytraits,
      title={Predicting the Big Five Personality Traits in Chinese Counselling Dialogues Using Large Language Models}, 
      author={Yang Yan and Lizhi Ma and Anqi Li and Jingsong Ma and Zhenzhong Lan},
      year={2024},
      eprint={2406.17287},
      archivePrefix={arXiv},
      primaryClass={cs.CL},
      url={https://arxiv.org/abs/2406.17287}, 
}

@article{amin2023will,
  title={Will affective computing emerge from foundation models and general artificial intelligence? A first evaluation of ChatGPT},
  author={Amin, Mostafa M and Cambria, Erik and Schuller, Bj{\"o}rn W},
  journal={IEEE Intelligent Systems},
  volume={38},
  number={2},
  pages={15--23},
  year={2023},
  publisher={IEEE}
}

@article{li2024automatic,
  title={Automatic evaluation for mental health counseling using llms},
  author={Li, Anqi and Lu, Yu and Song, Nirui and Zhang, Shuai and Ma, Lizhi and Lan, Zhenzhong},
  journal={arXiv preprint arXiv:2402.11958},
  year={2024}
}

@article{wang2024towards,
  title={Towards a Client-Centered Assessment of LLM Therapists by Client Simulation},
  author={Wang, Jiashuo and Xiao, Yang and Li, Yanran and Song, Changhe and Xu, Chunpu and Tan, Chenhao and Li, Wenjie},
  journal={arXiv preprint arXiv:2406.12266},
  year={2024}
}

@article{xu2024mentalLLM,
author = {Xu, Xuhai and Yao, Bingsheng and Dong, Yuanzhe and Gabriel, Saadia and Yu, Hong and Hendler, James and Ghassemi, Marzyeh and Dey, Anind K. and Wang, Dakuo},
title = {Mental-LLM: Leveraging Large Language Models for Mental Health Prediction via Online Text Data},
year = {2024},
issue_date = {March 2024},
publisher = {Association for Computing Machinery},
address = {New York, NY, USA},
volume = {8},
number = {1},
url = {https://doi.org/10.1145/3643540},
doi = {10.1145/3643540},
journal = {Proc. ACM Interact. Mob. Wearable Ubiquitous Technol.},
month = {mar},
articleno = {31},
numpages = {32}
}

@article{lamichhane2023evaluation,
  title={Evaluation of chatgpt for nlp-based mental health applications},
  author={Lamichhane, Bishal},
  journal={arXiv preprint arXiv:2303.15727},
  year={2023}
}

@inproceedings{lee-etal-2024-towards,
    title = "Towards Understanding Counseling Conversations: Domain Knowledge and Large Language Models",
    author = "Lee, Younghun  and
      Goldwasser, Dan  and
      Reese, Laura Schwab",
    editor = "Graham, Yvette  and
      Purver, Matthew",
    booktitle = "Findings of the Association for Computational Linguistics: EACL 2024",
    month = mar,
    year = "2024",
    address = "St. Julian{'}s, Malta",
    publisher = "Association for Computational Linguistics",
    url = "https://aclanthology.org/2024.findings-eacl.137",
    pages = "2032--2047",
}

@article{ji2023rethinking,
  title={Rethinking large language models in mental health applications},
  author={Ji, Shaoxiong and Zhang, Tianlin and Yang, Kailai and Ananiadou, Sophia and Cambria, Erik},
  journal={arXiv preprint arXiv:2311.11267},
  year={2023}
}

@misc{adhikary2024exploring,
      title={Exploring the Efficacy of Large Language Models in Summarizing Mental Health Counseling Sessions: A Benchmark Study}, 
      author={Prottay Kumar Adhikary and Aseem Srivastava and Shivani Kumar and Salam Michael Singh and Puneet Manuja and Jini K Gopinath and Vijay Krishnan and Swati Kedia and Koushik Sinha Deb and Tanmoy Chakraborty},
      year={2024},
      eprint={2402.19052},
      archivePrefix={arXiv},
      primaryClass={id='cs.CL' full_name='Computation and Language' is_active=True alt_name='cmp-lg' in_archive='cs' is_general=False description='Covers natural language processing. Roughly includes material in ACM Subject Class I.2.7. Note that work on artificial languages (programming languages, logics, formal systems) that does not explicitly address natural-language issues broadly construed (natural-language processing, computational linguistics, speech, text retrieval, etc.) is not appropriate for this area.'}
}

@article{hatcher2006development,
  title={Development and validation of a revised short version of the Working Alliance Inventory},
  author={Hatcher, Robert L and Gillaspy, J Arthur},
  journal={Psychotherapy research},
  volume={16},
  number={1},
  pages={12--25},
  year={2006},
  publisher={Taylor \& Francis}
}

@article{munder2010working,
  title={Working Alliance Inventory-Short Revised (WAI-SR): psychometric properties in outpatients and inpatients},
  author={Munder, Thomas and Wilmers, Fabian and Leonhart, Rainer and Linster, Hans Wolfgang and Barth, J{\"u}rgen},
  journal={Clinical Psychology \& Psychotherapy: An International Journal of Theory \& Practice},
  volume={17},
  number={3},
  pages={231--239},
  year={2010},
  publisher={Wiley Online Library}
}

@inproceedings{li-etal-2023-understanding,
    title = "Understanding Client Reactions in Online Mental Health Counseling",
    author = "Li, Anqi  and
      Ma, Lizhi  and
      Mei, Yaling  and
      He, Hongliang  and
      Zhang, Shuai  and
      Qiu, Huachuan  and
      Lan, Zhenzhong",
    editor = "Rogers, Anna  and
      Boyd-Graber, Jordan  and
      Okazaki, Naoaki",
    booktitle = "Proceedings of the 61st Annual Meeting of the Association for Computational Linguistics (Volume 1: Long Papers)",
    month = jul,
    year = "2023",
    address = "Toronto, Canada",
    publisher = "Association for Computational Linguistics",
    url = "https://aclanthology.org/2023.acl-long.577",
    doi = "10.18653/v1/2023.acl-long.577",
    pages = "10358--10376",
}

@misc{chiu2024computational,
      title={A Computational Framework for Behavioral Assessment of LLM Therapists}, 
      author={Yu Ying Chiu and Ashish Sharma and Inna Wanyin Lin and Tim Althoff},
      year={2024},
      eprint={2401.00820},
      archivePrefix={arXiv},
      primaryClass={cs.CL}
}

@article{horvath2011alliance,
  title={Alliance in individual psychotherapy},
  author={Horvath, Adam O and Del Re, AC and Fl{\"u}ckiger, Christoph and Symonds, Dianne},
  journal={Psychotherapy},
  volume={48},
  number={1},
  pages={9},
  year={2011},
  publisher={Educational Publishing Foundation}
}

@article{piper1991transference,
  title={Transference interpretations, therapeutic alliance, and outcome in short-term individual psychotherapy},
  author={Piper, William E and Azim, Hassan FA and Joyce, Anthony S and McCallum, Mary},
  journal={Archives of general Psychiatry},
  volume={48},
  number={10},
  pages={946--953},
  year={1991},
  publisher={American Medical Association}
}

@article{horvath1991relation,
  title={Relation between working alliance and outcome in psychotherapy: A meta-analysis.},
  author={Horvath, Adam O and Symonds, B Dianne},
  journal={Journal of counseling psychology},
  volume={38},
  number={2},
  pages={139},
  year={1991},
  publisher={American Psychological Association}
}

@article{bachelor2000participants,
  title={Participants' perceptions of dimensions of the therapeutic alliance over the course of therapy},
  author={Bachelor, Alexandra and Salam{\'e}, Ramzi},
  journal={The Journal of psychotherapy practice and research},
  volume={9},
  number={1},
  pages={39},
  year={2000},
  publisher={American Psychiatric Publishing}
}

@inproceedings{li-etal-2023-coannotating,
    title = "{C}o{A}nnotating: Uncertainty-Guided Work Allocation between Human and Large Language Models for Data Annotation",
    author = "Li, Minzhi  and
      Shi, Taiwei  and
      Ziems, Caleb  and
      Kan, Min-Yen  and
      Chen, Nancy  and
      Liu, Zhengyuan  and
      Yang, Diyi",
    editor = "Bouamor, Houda  and
      Pino, Juan  and
      Bali, Kalika",
    booktitle = "Proceedings of the 2023 Conference on Empirical Methods in Natural Language Processing",
    month = dec,
    year = "2023",
    address = "Singapore",
    publisher = "Association for Computational Linguistics",
    url = "https://aclanthology.org/2023.emnlp-main.92/",
    doi = "10.18653/v1/2023.emnlp-main.92",
    pages = "1487--1505"
}

@article{monroe2008fightin,
  title={Fightin'words: Lexical feature selection and evaluation for identifying the content of political conflict},
  author={Monroe, Burt L and Colaresi, Michael P and Quinn, Kevin M},
  journal={Political Analysis},
  volume={16},
  number={4},
  pages={372--403},
  year={2008},
  publisher={Cambridge University Press}
}

@article{lalk2024alliance,
author = {Lalk, Christopher and Steinbrenner, Tobias and Kania, Weronika and Popko, Alexander and Wester, Robin and Schaffrath, Jana and Eberhardt, Steffen and Schwartz, Brian and Lutz, Wolfgang and Rubel, Julian},
year = {2024},
month = {03},
pages = {1-16},
title = {Measuring Alliance and Symptom Severity in Psychotherapy Transcripts Using Bert Topic Modeling},
volume = {51},
journal = {Administration and Policy in Mental Health and Mental Health Services Research},
doi = {10.1007/s10488-024-01356-4}
}

@misc{deepseekai2025deepseekr1,
      title={DeepSeek-R1: Incentivizing Reasoning Capability in LLMs via Reinforcement Learning}, 
      author={DeepSeek-AI},
      year={2025},
      eprint={2501.12948},
      archivePrefix={arXiv},
      primaryClass={cs.CL},
      url={https://arxiv.org/abs/2501.12948}, 
}

@article{qwen2.5,
    title   = {Qwen2.5 Technical Report}, 
    author  = {An Yang and Baosong Yang and Beichen Zhang and Binyuan Hui and Bo Zheng and Bowen Yu and Chengyuan Li and Dayiheng Liu and Fei Huang and Haoran Wei and Huan Lin and Jian Yang and Jianhong Tu and Jianwei Zhang and Jianxin Yang and Jiaxi Yang and Jingren Zhou and Junyang Lin and Kai Dang and Keming Lu and Keqin Bao and Kexin Yang and Le Yu and Mei Li and Mingfeng Xue and Pei Zhang and Qin Zhu and Rui Men and Runji Lin and Tianhao Li and Tingyu Xia and Xingzhang Ren and Xuancheng Ren and Yang Fan and Yang Su and Yichang Zhang and Yu Wan and Yuqiong Liu and Zeyu Cui and Zhenru Zhang and Zihan Qiu},
    journal = {arXiv preprint arXiv:2412.15115},
    year    = {2024}
}

@article{zheng2024llamafactory,
  title={LlamaFactory: Unified Efficient Fine-Tuning of 100+ Language Models}, 
  author={Yaowei Zheng and Richong Zhang and Junhao Zhang and Yanhan Ye and Zheyan Luo and Yongqiang Ma},
  journal={arXiv preprint arXiv:2403.13372},
  year={2024},
  url={http://arxiv.org/abs/2403.13372}
}

@inproceedings{lin-etal-2024-working,
    title = "Working Alliance Transformer for Psychotherapy Dialogue Classification",
    author = "Lin, Baihan  and
      Cecchi, Guillermo  and
      Bouneffouf, Djallel",
    editor = "Naumann, Tristan  and
      Ben Abacha, Asma  and
      Bethard, Steven  and
      Roberts, Kirk  and
      Bitterman, Danielle",
    booktitle = "Proceedings of the 6th Clinical Natural Language Processing Workshop",
    month = jun,
    year = "2024",
    address = "Mexico City, Mexico",
    publisher = "Association for Computational Linguistics",
    url = "https://aclanthology.org/2024.clinicalnlp-1.6/",
    doi = "10.18653/v1/2024.clinicalnlp-1.6",
    pages = "64--69"
}

@article{lin2025compass,
  title={COMPASS: Computational mapping of patient-therapist alliance strategies with language modeling},
  author={Lin, Baihan and Bouneffouf, Djallel and Landa, Yulia and Jespersen, Rachel and Corcoran, Cheryl and Cecchi, Guillermo},
  journal={Translational Psychiatry},
  volume={15},
  number={1},
  pages={166},
  year={2025},
  publisher={Nature Publishing Group UK London}
}

\appendix
\section{Data Analysis}
\label{appendix: data_analysis}





\begin{figure}
    \centering
    \scalebox{1.9}{
    \includegraphics[width=0.5\linewidth]{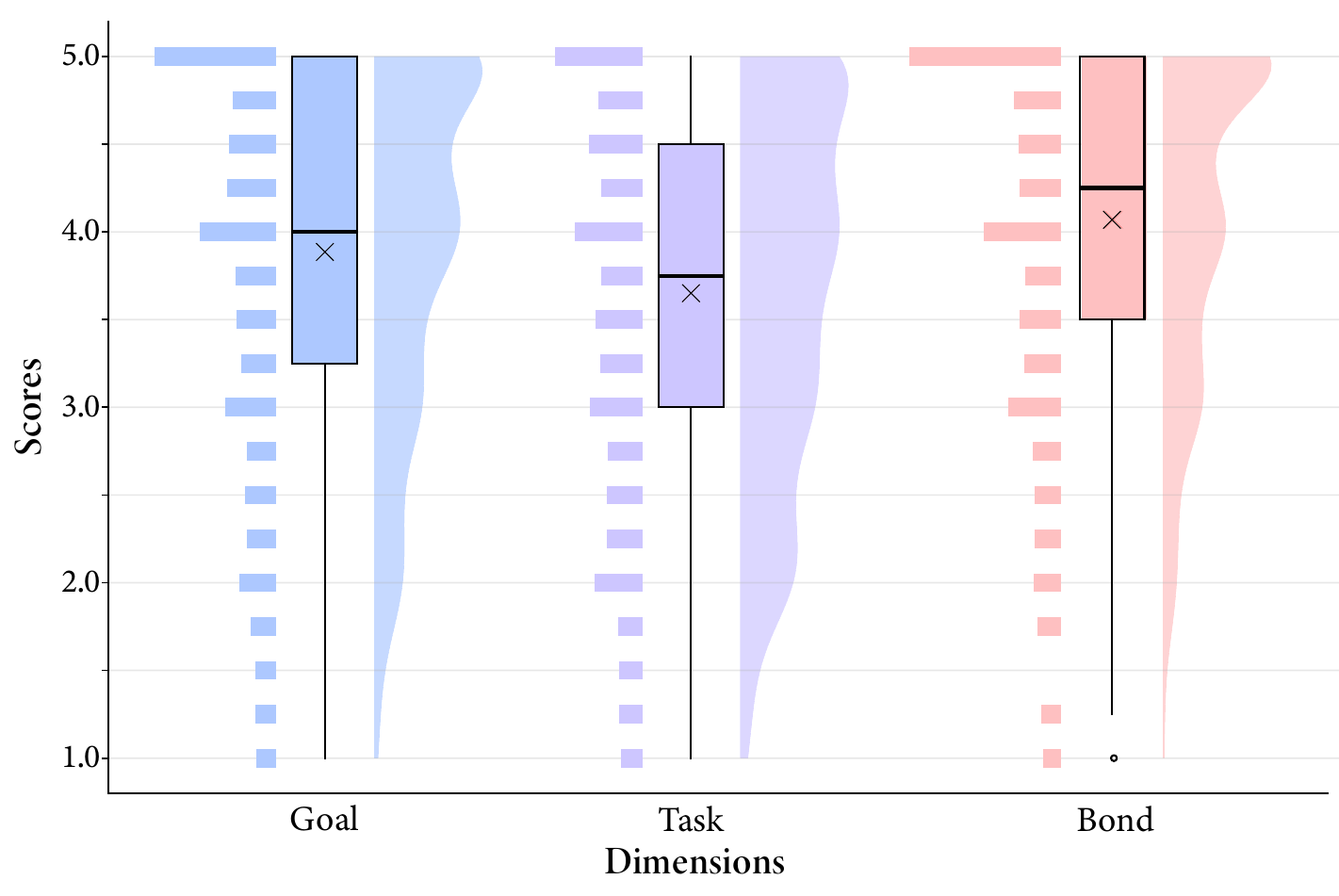}}
    \caption{Distributions of client ratings across the three working alliance dimensions—Goal, Task, and Bond—illustrated using histograms, box plots, and half-violin plots. Within each box, the X denotes the mean. The central line represents the median, the box corresponds to the interquartile range (IQR, 25th–75th percentile), and the whiskers extend to the full range of the data.}
    \label{fig:client_wai_distribution}
\end{figure}


Figure~\ref{fig:client_wai_distribution} shows the score distribution of the CounselingWAI dataset. The score distributions across the three dimensions exhibit a typical negative skew, a pattern commonly observed in real-world alliance assessments~\citep{tryon2008magnitude, goldberg2020machine}. This characteristic indicates that the dataset used for subsequent model training and evaluation reflects authentic client rating patterns. The average scores for the \textit{Goal}, \textit{Task} and \textit{Bond} dimensions all surpass 3.5, suggesting that a relatively robust therapeutic relationship can be established between counselors and clients in online text-based psychological counseling.

In addition, we analyzed the crafted rationales. The generated rationales had an overall average length of 155.25 characters, with a standard deviation of 13.27. To examine lexical patterns, we extracted features for each dimension's rationales by computing the log odds ratio with an informative Dirichlet prior~\citep{monroe2008fightin}, comparing all unigrams in the rationales of one dimension against the other two. The top five phrases for each dimension are presented in Table~\ref{tab:data_rationales}. Rationales for each dimension were significantly associated (z-score > 3) with specific key phrases (e.g., \textit{Goal} with "establish," \textit{Task} with "method," \textit{Bond} with "support").

\begin{table}[]
\scalebox{0.7}{
\begin{tabular}{lll}
\toprule
\textbf{Dim} & \textbf{Length} & \textbf{Lexical Features}                                                                                                                           \\
\midrule
Goal               & $156.71_{13.66}$     & \begin{tabular}[l]{@{}l@{}}目标/goal (55.32), 努力/effort (29.31), \\ 探讨/explore (27.57), 改善/improve (24.97), \\ 制定/establish (24.38)\end{tabular}       \\
\midrule
Task               & $151.76_{12.80}$     & \begin{tabular}[l]{@{}l@{}}方法/method (41.73), 意识/aware (40.05), \\ 新/new (37.80), 改变/change (29.41) \\ 解决问题/problem-solving (29.70)\end{tabular}   \\
\midrule
Bond               & $157.29_{13.54}$     & \begin{tabular}[l]{@{}l@{}}支持/support (45.51), 关心/care (38.76), \\ 表现/express (37.59), 感受/feeling (37.05), \\ 理解/understand (36.54)\end{tabular}
\\
\bottomrule
\end{tabular}}
\caption{Data characteristics of generated rationales for each working alliance dimension, including the mean with standard deviation of rationale length and key lexical features. The rightmost column presents the five words most strongly associated with each dimension's rationales, with rounded z-scored log-odds ratios provided in parentheses.}
\label{tab:data_rationales}
\end{table}




\section{Background of Therapeutic Alliance}

The therapeutic alliance is widely recognized as a foundational construct in psychotherapy and is commonly conceptualized as comprising three interrelated components: \textbf{Goal}, \textbf{Task}, and \textbf{Bond}~\citep{bordin1979generalizability}.

\textbf{Goal.} Establishing clear counseling goals is fundamental to a successful counseling session, distinguishing it from casual conversations. Therapeutic goals involve fostering positive changes in clients' thoughts, cognition, and behaviors, facilitated by counselors’ guidance and support. Both counselors and clients should collaboratively define and mutually agree on their counseling goals, ensuring their efforts are directed toward shared objectives.

\textbf{Task.} Beyond setting consistent goals, reaching mutual agreement between counselors and clients on specific methods to achieve them is a critical element. Counselors typically propose tasks based on their personal styles, experience, and predispositions, but clients may find them unmanageable or unsuitable. In such instances, counselors need to provide alternative approaches to better engage their clients. Furthermore, counselors should clarify how these tasks contribute to achieving therapy goals, as this understanding is crucial for effective treatment~\citep{horvath1993role}.

\textbf{Bond.} In addition to the cognitive aspects of the alliance that emphasize the consensus on therapy goals and tasks, the emotional attachment between counselors and clients is crucial. The bond reflects the feelings and attitudes that each party holds toward the other, fostering collaboration and trust. When clients perceive counselors' genuine care and attention, they feel secure and motivated to engage in therapy. Likewise, when both parties trust each other's abilities, a shared commitment to goals and tasks can be established.

\section{Automatic Prediction}
\subsection{Template Prompt}
\label{template_prompt}
\begin{figure}
    \centering
    \scalebox{1.95}{
    \includegraphics[width=0.5\linewidth]{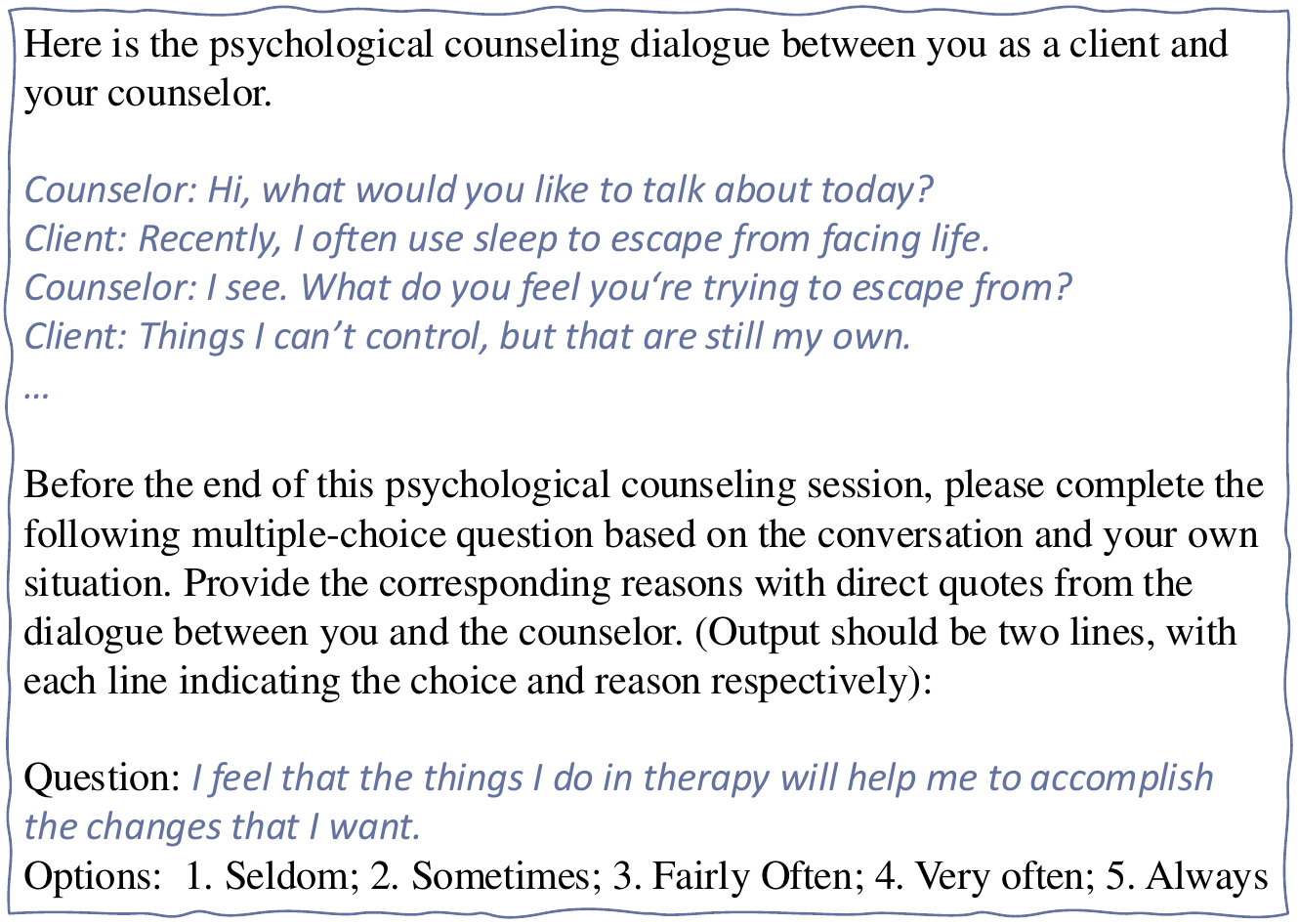}}
    \caption{The template prompt for instructing LLMs to predict clients' perceived working alliance, using an example conversation and questionnaire question (displayed in \textcolor{CadetBlue}{\textit{cadetblue italic}} text).}
    \label{fig:prompt}
\end{figure}

The template prompt used to instruct LLMs to predict clients' perceived working alliance is shown in Figure~\ref{fig:prompt}.

\subsection{Experimental Settings}
\label{appendix: experimental_settings}
Table~\ref{tab:hyperparameters} shows the key hyperparameters and corresponding values used in our fine-tuning experiments.

\subsection{Experimental Results}

Table~\ref{tab:case} presents the example cases where the model's predictions significantly deviate from the client's self-reported scores.

\begin{table*}[]
\centering
\scalebox{0.9}{
\begin{tabular}{ll}

\textbf{Conversation}                             & \textbf{Evaluation Results}                \\ \hline

\makecell[l]{Counselor: Hello, are you online?\\ \textit{10 minutes later...}\\ Client: Sorry, I fell asleep.\\ Counselor: It's okay, let's begin now.\\ Client: Thanks for waiting for me.\\ \textit{Counseling in progress...}\\ Counselor: Do you have any plans for progress?\\ Counselor: Are you still online? Hello?}   & \makecell[l]{\textbf{Dimension}: Bond\\ \textbf{Client}: 1.25\\ \textbf{CARE}: 4.25\\ \textbf{Explanations}: The counselor patiently waits for \\ the client's responses, even when the client falls \\ asleep, without showing any impatience. The \\ client also responds in a friendly manner, such as \\ saying "Sorry" and "Thanks". These interactions \\ reflect mutual respect and understanding...}  \\ \hline

\makecell[l]{\textit{Counseling in progress...}\\ Client: I’ve been doing counseling for a while, \\ and slowly I’ve started to shift my focus. I’m feeling \\ a bit better now. \\\textit{Counseling in progress...}\\ Counselor: You don’t want to have a negative \\ impact on her?\\ Client: Yes.\\ Counselor: You don’t want to put her in a diff-\\icult position?\\ Client: I guess so. \\ \textit{Counseling in progress...}}  & \makecell[l]{\textbf{Dimension}: Task\\ \textbf{Client}: 1.0\\ \textbf{CARE}: 4.0\\ \textbf{Explanations}: The client directly stated that \\ counseling has helped them clarify the areas \\ that need change, and feel better by shifting \\ focus...} \\ \hline
\end{tabular}}
\caption{Example cases where the model's predictions significantly deviate from the client's self-reported scores.}
\label{tab:case}
\end{table*}






\section{Additional LLM-based Insights}

\subsection{Interaction Patterns and Therapeutic Alliance}

Table~\ref{tab:interaction_wai} shows the regression coefficients for counselor–client interaction patterns predicting session-level working alliance scores across the three dimensions.

\subsection{Example Explanations}
\label{example_explanations}

Table~\ref{tab:explanations_example} presents some example explanations generated by our best-performing model.

\begin{table}[]
\centering
\scalebox{0.8}{
\begin{tabular}{lc}
\hline
\textbf{Hyperparameters}    & \textbf{Value} \\
\hline
Per-device Train Batchsize  & 1              \\
Gradient Accumulation Steps & 2              \\
Warmup Ratio                & 0.1            \\
LR Scheduler Type           & cosine         \\
Learning Rate               & 5e-7           \\
Data Type                   & bfloat16       \\
Optimizer                   & adamw          \\
Epoch                       & 10       \\ \hline
\end{tabular}}
\caption{The hyperparameters with values used in our fine-tuning experiments.}
\label{tab:hyperparameters}
\end{table}

\begin{table}[]
\centering
\scalebox{0.8}{
\begin{tabular}{llll}
\toprule
                       & \textbf{Goal}              & \textbf{Task}              & \textbf{Bond}              \\
\midrule
Supporting - Positive  & 0.17              & 0.84              & 0.62              \\
Challenging - Positive & \textbf{2.07***}  & \textbf{2.64***}  & \textbf{1.47***}  \\
Supporting - Negative  & \textbf{-5.71***} & \textbf{-5.57***} & \textbf{-6.36***} \\
Challenging - Negative & -1.70           & -2.74            & -1.77   \\
\bottomrule
\end{tabular}}
\caption{Regression coefficients for counselor–client interaction patterns predicting session-level working alliance scores across the three dimensions; *** denotes p < 0.001.}
\label{tab:interaction_wai}
\end{table}

\begin{table*}[]
\centering
\scalebox{0.9}{
\begin{tabular}{ccl}
\hline
\textbf{Dimension}        & \textbf{Evaluation} & \multicolumn{1}{c}{\textbf{Example}}                                                                                                                                                                                                                                                                                                                                                                                                                                                                                                                                                                                                                                                                                                                                                              \\
\hline
\multirow{2}{*}{Goal}     & bad                 & \begin{tabular}[c]{@{}l@{}}In our conversations, although the counselor tried to guide me in exploring issues \\ and feelings, we did not clearly agree on specific areas for change. For example, \\ when I mentioned how to deal with a difficult roommate, the counselor focused \\ more on understanding my feelings and situation rather than collaboratively \\ determining specific change strategies. My responses, such as “No thanks” and \\ later mentioning “I might not be suited for this,” indicate that my understanding of \\ the counseling process may differ from the counselor's expectations. As a result, \\ we rarely reached a consensus on the areas where I needed to make efforts for \\ change.\end{tabular}                                                         \\ \cline{2-3}
                          & good                & \begin{tabular}[c]{@{}l@{}}In the conversation, the counselor and I jointly established counseling goals, such \\ as, “Let’s set the goal to increase daily sleep time to 6 hours.” This indicates that \\ we were indeed actively involved in the goal-setting process, which was a \\ continuous part of the counseling sessions, thus aligning with the description of \\ “always” collaboratively setting goals.\end{tabular}                 \\ \hline
\multirow{2}{*}{Approach} & bad                 & \begin{tabular}[c]{@{}l@{}}In the conversation, I expressed my feelings about the counselor's questions, \\ saying, “To be honest, I speak directly and bluntly, and this is a free service. I \\ shouldn’t have any expectations. I really shouldn’t have too many demands and \\ should be content.” This indicates that I did not feel the counseling provided new \\ perspectives or methods for addressing my relationship issues. The counselor \\ focused more on listening and understanding, whereas I was expecting concrete \\ guidance and solutions, such as, “Shouldn’t you be better and have more ideas?” \\ This shows that I didn’t frequently find new ways to view my problems during the \\ counseling sessions.\end{tabular}                                                \\ \cline{2-3}
                          & good                & \begin{tabular}[c]{@{}l@{}}In the conversation, I mentioned trying to make changes but not finding the right \\ approach, which indicates that I recognize the value of the counseling process. The \\ counselor helped me reflect on my concerns about effort and boundaries, for \\ example, “It seems like you feel that the evaluation of your goodness comes from \\ what you do rather than your traits or personal charisma in relationships.” This \\ helped me start to understand my behavioral patterns more deeply. Although I am \\ still exploring, I feel that counseling has indeed guided me toward the changes I \\ want.\end{tabular}                                                                                                                                          \\
\hline
\multirow{2}{*}{Bond}     & bad                 & \begin{tabular}[c]{@{}l@{}}In the conversation, I expressed doubts about the counselor's professionalism and \\ mentioned, “You gave me the impression of being a bit rushed, not allowing me time \\ to process my changes,” as well as “I have a lot of anger towards you.” This \\ indicates that I did not feel that the counselor was attuned to my personal \\ preferences but rather sensed a mismatch in professional approach.\end{tabular}                                                                                          \\ \cline{2-3}
& good                & \begin{tabular}[c]{@{}l@{}}During the counseling process, the counselor demonstrated patience and \\ understanding, such as by asking, “Can you describe the sense of presence you \\ have experienced from childhood to now?” This indicates that the counselor \\ actively listens and is concerned about my feelings. Although personal preferences \\ were not directly addressed, the counselor's positive feedback and in-depth \\ exploration, such as, “The sense of presence you described—when you feel that \\ doing something genuinely brings joy to others—seems to be part of your value \\ system. It looks like you have a direction, but the path is blocked,” show the \\ counselor’s deep understanding and care for me, making me feel valued and \\ supported.\end{tabular} \\
\hline
\end{tabular}}
\caption{Example explanations generated by our best-performing model.}
\label{tab:explanations_example}
\end{table*}




\section{Limitations}

First, while our approach is validated on text-based counseling, its principles are applicable to other modalities (e.g., face-to-face or video sessions) through speech-to-text conversion. Future work could extend this by integrating multi-modal signals such as vocal tone and facial expressions.

Second, although the data comes from a single Chinese platform, the underlying mechanisms of therapeutic alliance are largely universal, and our diverse dataset supports cross-context generalizability. Future efforts will test and adapt the model across languages and cultures through fine-tuning and expanded data collection. Additionally, the lack of strictly longitudinal data limits our ability to assess temporal dynamics, a direction we plan to address in subsequent studies.

Third, the rationales used to augment dialogues are derived from expert annotations rather than client self-reports, given the practical challenges in obtaining scalable first-person explanations. While this aligns with clinical supervision practices, it may introduce professional bias, meaning the model learns expert inference patterns rather than client-internal reasoning. Future work should explore hybrid approaches combining expert and client perspectives.

Finally, there remains room to improve the model’s correlation with client ratings and the precision of rationale generation. We will refine training strategies, optimize instructions, and expand datasets to enhance performance and robustness. Despite these limitations, this work offers meaningful insights into the integration of NLP and psychological process research.

\section{Human Data Collection}
The newly collected human data in this study consist of expert-authored explanatory rationales augmenting pairs of original counseling dialogues and corresponding client-rated working alliance inventory items.

(1) For each dialogue, experts were provided with the full transcript, the relevant working alliance item, and the client's rating. Drawing on their clinical and supervisory expertise, they identified evidence from the dialogue and synthesized it into a structured rationale explicitly linking textual evidence to the alliance score.

(2) Experts were recruited from universities and professional institutions. All held valid counseling licenses and possessed formal supervisory qualifications and substantial practice experience. Each expert received approximately 10k RMB as compensation for their annotation work.

(3) All participants provided written informed consent prior to participation. They were fully informed of the study procedures, research objectives, data usage policies, and their right to withdraw at any time without penalty.

(4) This study was reviewed and approved by the appropriate Institutional Review Board (IRB).

\section{Ethics Statement}


\paragraph{Data Privacy and Release.} The original counseling dataset is fully anonymized and released under strictly controlled research conditions. Access requires a formal application process, which includes submission of valid identification, a detailed research justification, proof of full-time academic affiliation, and documented IRB approval from the applicant's institution. Only full-time principal investigators are eligible to apply, and each application is subject to review by the applicant's Office of Research. Approved applicants must also sign a legally binding Data Non-Disclosure Agreement, committing not to share the data with any third party. The augmented dataset is available under identical terms, and eligibility is contingent upon prior access to the original dataset.

\paragraph{LLM-based Predictions.} This study aims to introduce an automated approach for predicting client-rated therapeutic alliance with their counselors in the context of online text-based counseling. We advocate for using LLM-based predictions as an additional tool to help counselors gain a deeper understanding of their clients. It is \textit{not} designed for clinical diagnosis, therapeutic intervention, or as a replacement for professional clinical judgment. Given the inherent limitations of current LLMs, predictions should be interpreted with caution. Furthermore, we highlight the importance of considering broader societal and ethical implications, including safeguards against potential misuse, when implementing such technology in sensitive domains.


\end{CJK*}
\end{document}